\documentclass[letterpaper, 10 pt, conference]{ieeeconf}  

\IEEEoverridecommandlockouts                              

\usepackage{graphicx}
\usepackage{epsfig}
\usepackage{epic,eepic}
\usepackage{times}
\usepackage{mathptmx}
\usepackage{amsfonts}
\usepackage{amsmath}
\usepackage{cite}

\usepackage[dvipdfm]{color}

\usepackage{times}
\usepackage{multirow}

\definecolor{red}{rgb}{1,0,0}
\definecolor{green}{rgb}{0,1,0}
\definecolor{blue}{rgb}{0,0,1}
\definecolor{violet}{rgb}{1,0,1}
\definecolor{cyan}{cmyk}{1,0,0,0}
\definecolor{magenta}{cmyk}{0,1,0,0}
\definecolor{yellow}{cmyk}{0,0,1,0}
\definecolor{white}{rgb}{1,1,1}

\newcommand{\CommentOut}[1]{}

\newcommand{\FIG}[3]{
\begin{minipage}[b]{#1cm}
\begin{center}
\includegraphics[width=#1cm]{#2}
{\scriptsize #3}
\end{center}
\end{minipage}
}

\newcommand{\FIGRM}[4]{
\begin{minipage}[b]{#1cm}
\begin{center}
\includegraphics[angle=-90,clip,width=#1cm]{#2}\vspace*{1mm}
\\
{\scriptsize #3}
\vspace*{#4mm}
\end{center}
\end{minipage}
}

\begin{document}

\newcommand{\figA}{
\begin{figure}[t]
\begin{center}
\begin{center}
\FIG{8.5}{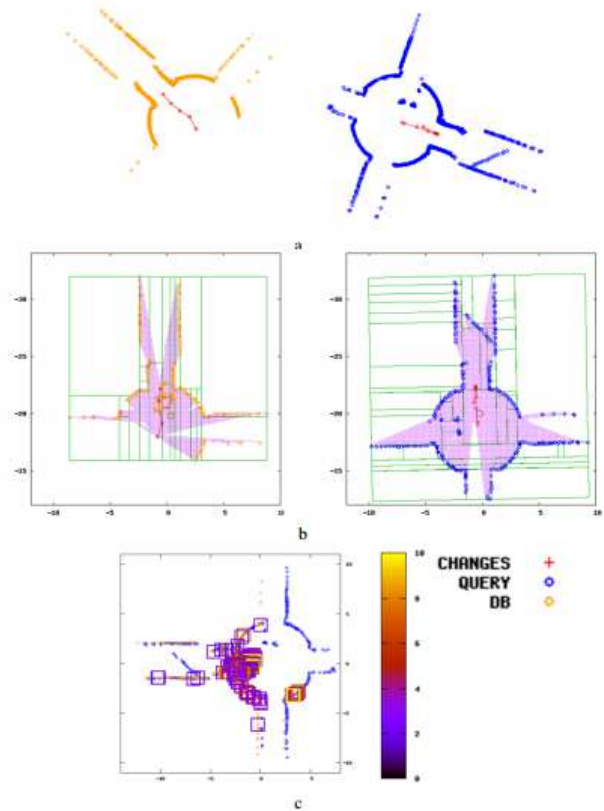}{}
\caption{The key idea is viewpoint planning in which the origin of the local map coordinate (termed ``viewpoint") is 
planned by scene parsing 
and determined by our ``viewpoint planner"
to be invariant against small variations in self-location and changes,
which 
aims at providing similar viewpoints for similar scenes (i.e., the relevant map pair) 
and enables a direct comparison of 
both the appearance and the pose of visual features 
between each map pair
(i.e., without requiring pre-alignment of each map pair).
(a) 
A query local map (left) and a database local map (right)
together with the robot's trajectory (red points).
(b)
Scene parsing results (green line segments)
and planned viewpoint (the big red point).
(c)
Detected anomaly (small colored boxes)
and anomaly-ness score (color bar).
}\label{fig:A}
\end{center}
\vspace*{-8mm}
\end{center}
\end{figure}
}

\newcommand{\figGd}{
\begin{figure*}[t]
\begin{center}
\FIG{17}{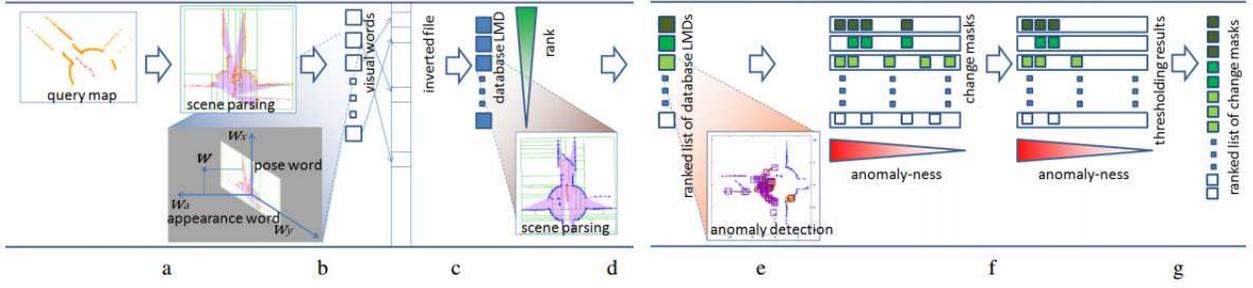}{}
\end{center}
\caption{The overall pipeline of the algorithm, 
which involves
four main steps:
viewpoint planning (a),
local map descriptor (b),
global self-localization (c: database retrieval, d: SPM matching)
and 
change detection (e: anomaly detection, f: thresholding, g: re-ranking),
which 
are described in sections \ref{sec:app}, \ref{sec:lmd}, \ref{sec:loc}, and \ref{sec:cd}, respectively.
}\label{fig:G}
\vspace*{-5mm}
\end{figure*}
}

\newcommand{\hs}{\hspace*{-2mm}}

\newcommand{\figB}{
\begin{figure*}[t]
\begin{center}
\begin{center}
\FIG{17}{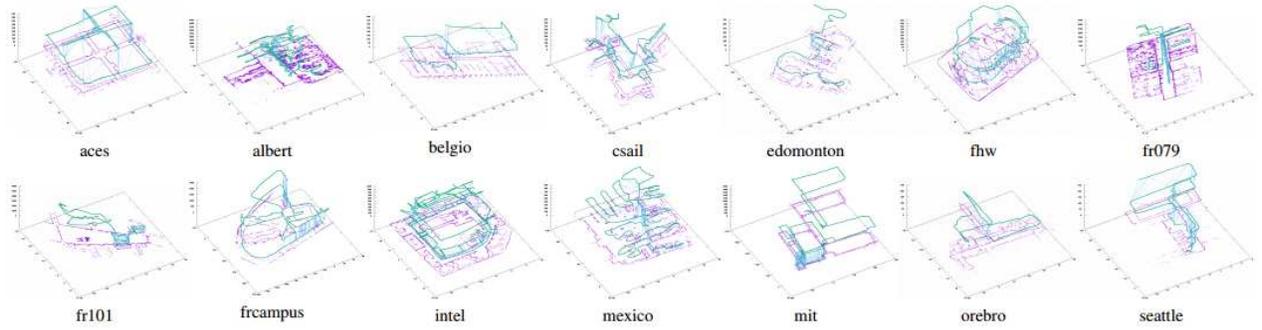}{}
\caption{Datasets. 
(Purple points: point clouds. 
Green curve: robot's trajectory.
Each of the light blue line segments connects 
a local map pair
that corresponds to each ground-truth loop closing.)}\label{fig:B}
\end{center}
\vspace*{-5mm}
\end{center}
\end{figure*}
}

\newcommand{\figF}{
\begin{figure*}[t]
\begin{center}
\begin{center}
\FIG{17}{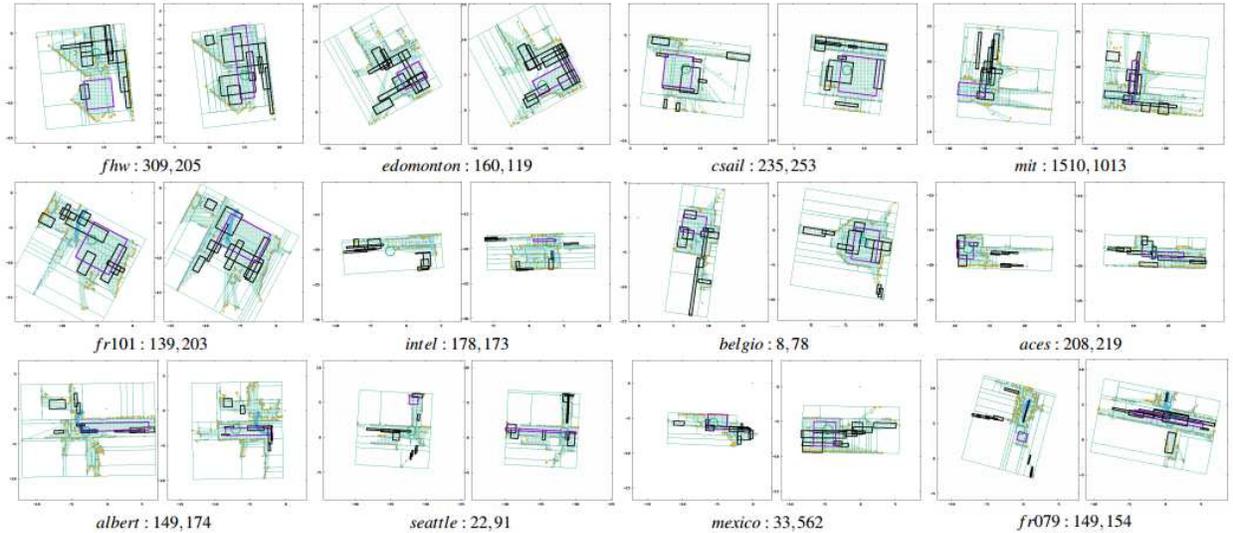}{}
\caption{
Samples of scene parsing. 
12 ($=3\times 4$) different 
pairs of scene parsing are shown 
for 12
relevant pairings of 
a query local map (left) 
and a relevant database local map (right) with 
``query map's ID, database map's ID". 
(Orange points: datapoints 
from the original local map. 
Boxes: ``room" primitives proposed by the CoR method. 
The green big circle: planned viewpoint. 
Small blue points with lines: the robot's trajectory. 
Green dots: unoccupied cells.)}\label{fig:F}
\end{center}
\vspace*{-7mm}
\end{center}
\end{figure*}
}

\newcommand{\figC}{
\begin{figure*}[t]
\begin{center}
\begin{center}
\FIG{17}{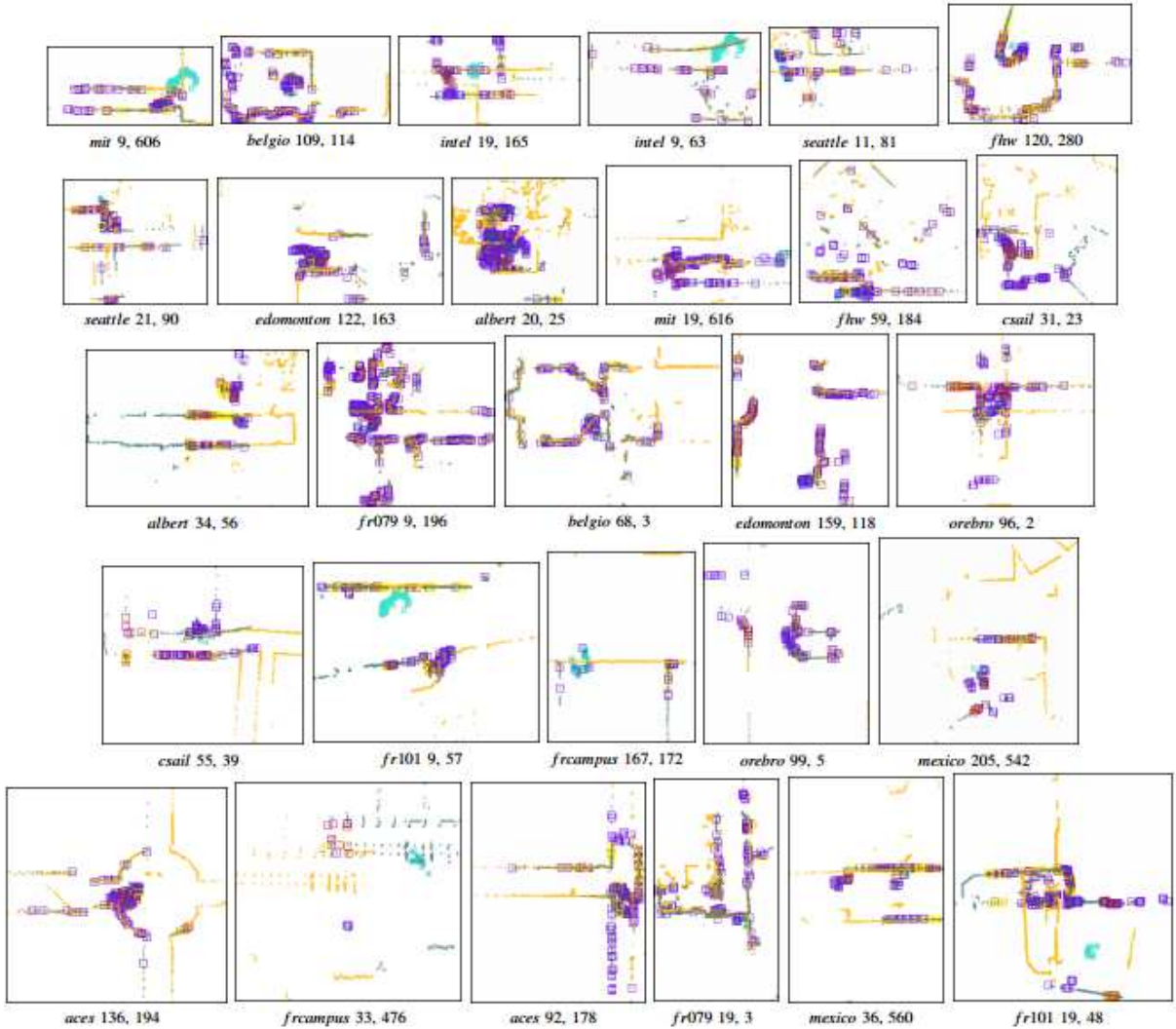}{}
\caption{Change detection. 
Point clouds of the query and relevant database maps 
and ground-truth changes (light blue points) 
are overlaid using the information of the 
{\it planned} viewpoint, 
and are shown with ``query map's ID, database map's ID". 
The meaning of the datapoints, anomalies, 
and anomaly-ness is the same as in Fig.\ref{fig:A}.}\label{fig:C}
\end{center}
\vspace*{-6mm}
\end{center}
\end{figure*}
}

\newcommand{\figD}{
\begin{figure*}[t]
\begin{center}
\begin{center}
\FIG{17}{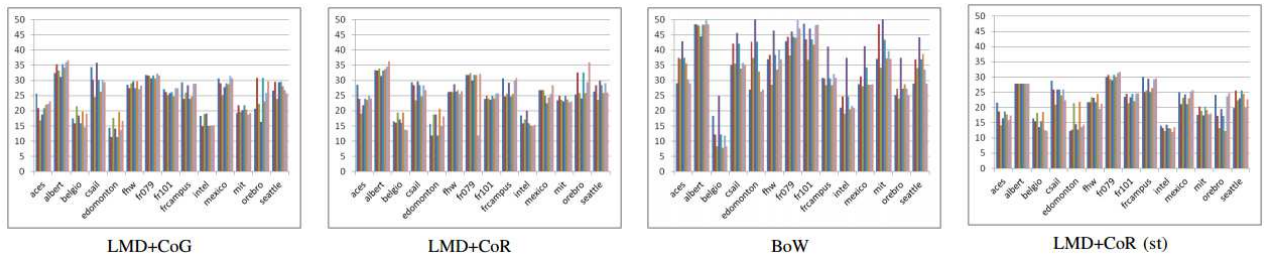}{}
\caption{Self-localization performance. (Vertical axis: ANR[\%]. For each dataset, from left to right, results for descriptor \#1, ..., \#8 are shown. )}\label{fig:D}
\end{center}
\vspace*{-2mm}
\end{center}
\end{figure*}
}

\newcommand{\figE}{
\begin{figure*}[t]
\begin{center}
\vspace*{-5mm}
\begin{center}
\hspace*{-0.5cm}\FIGRM{3.4}{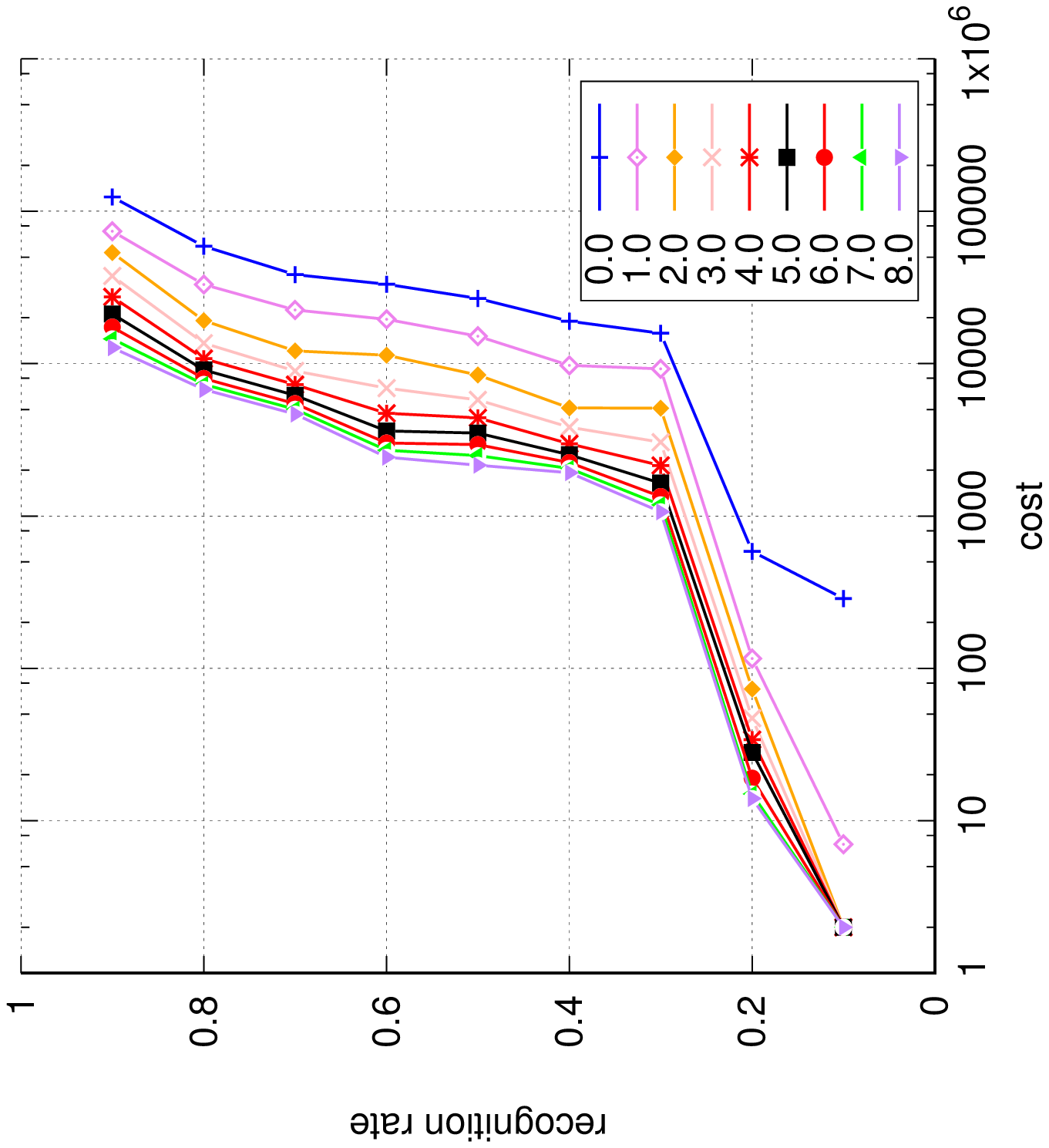}{$mit$}{1.5}%
\hspace*{-1.2cm}\FIGRM{3.4}{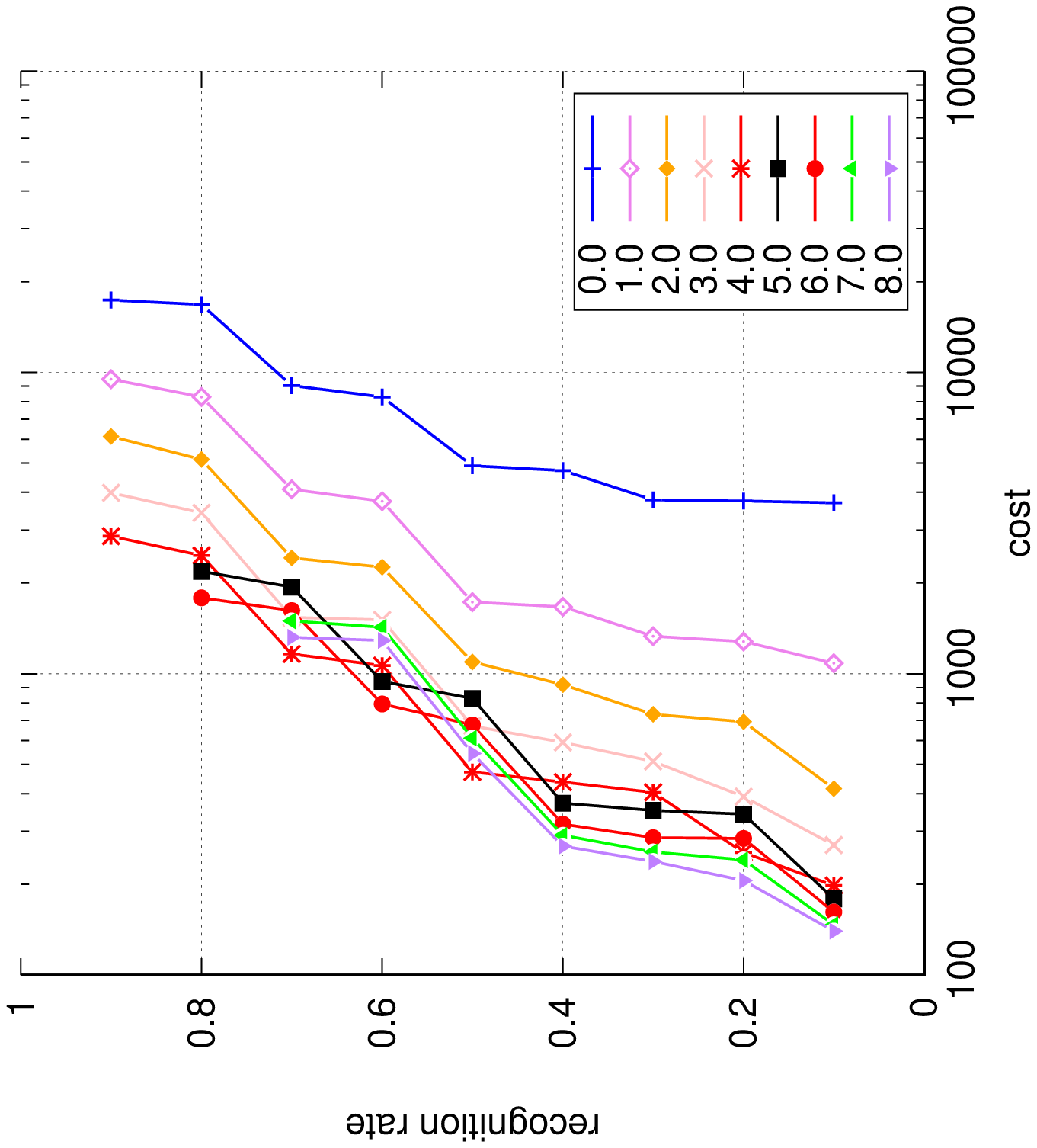}{$belgio$}{1}%
\hspace*{-1.2cm}\FIGRM{3.4}{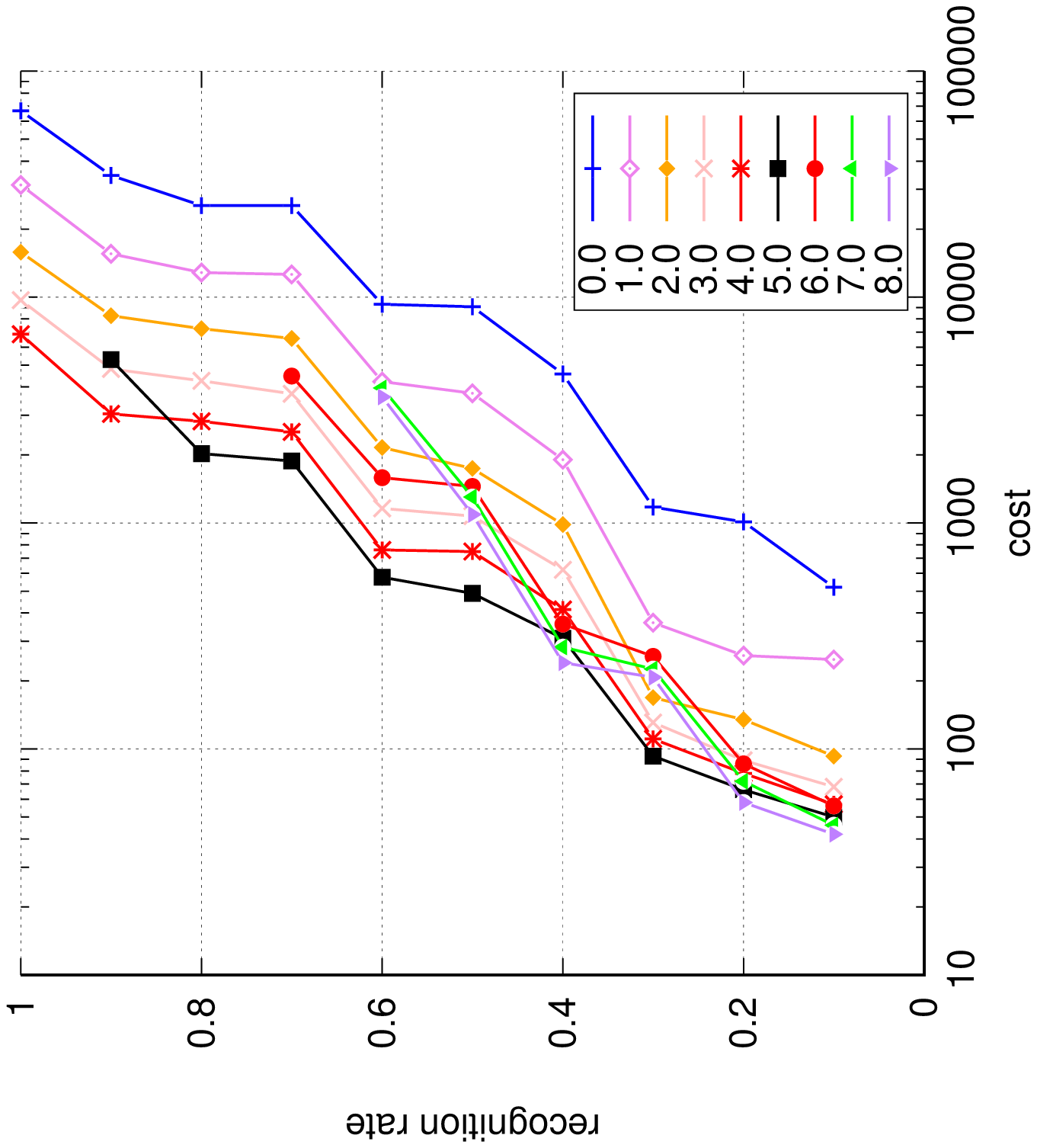}{$intel$}{1.5}%
\hspace*{-1.2cm}\FIGRM{3.4}{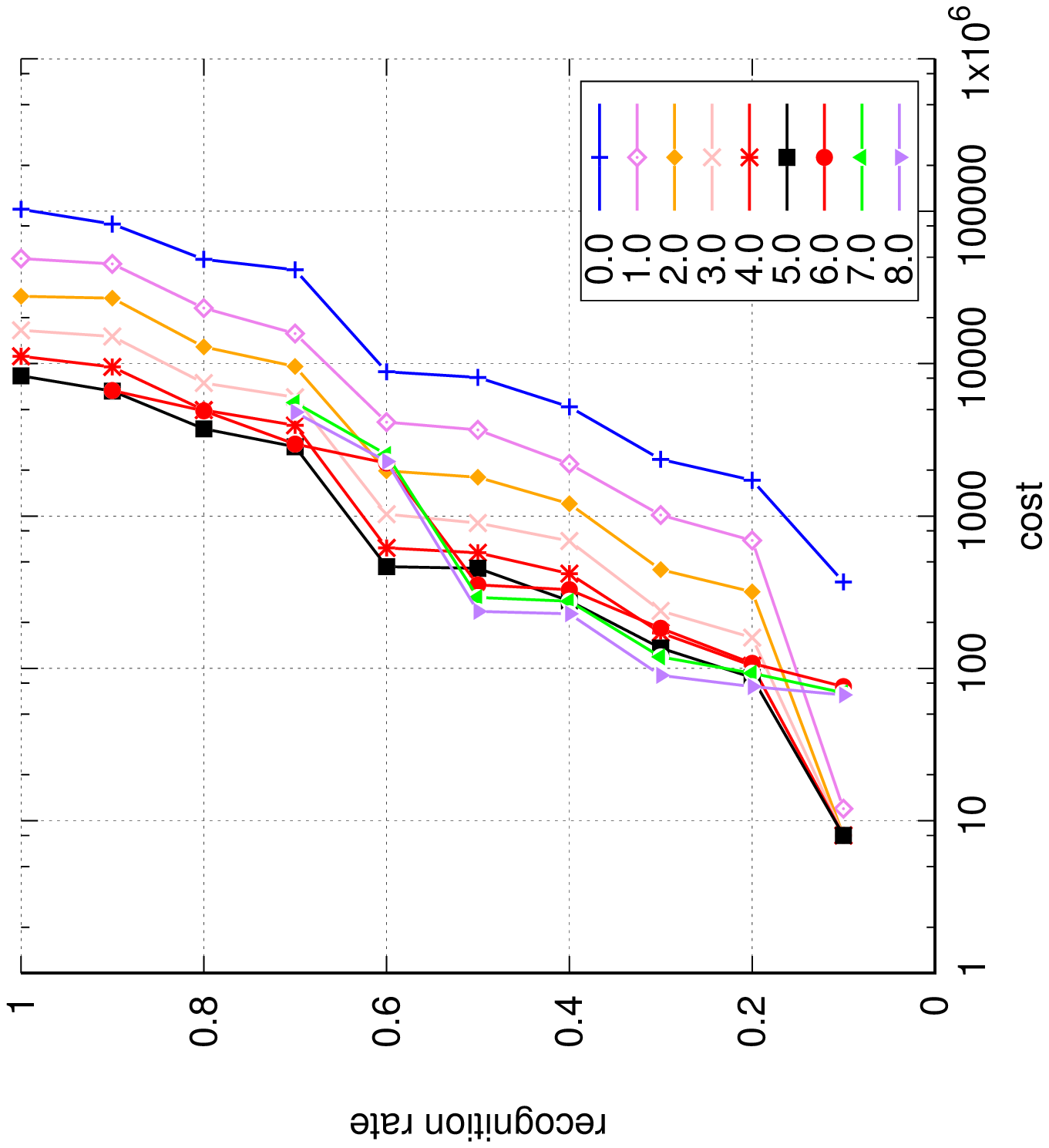}{$fhw$}{1}%
\hspace*{-1.2cm}\FIGRM{3.4}{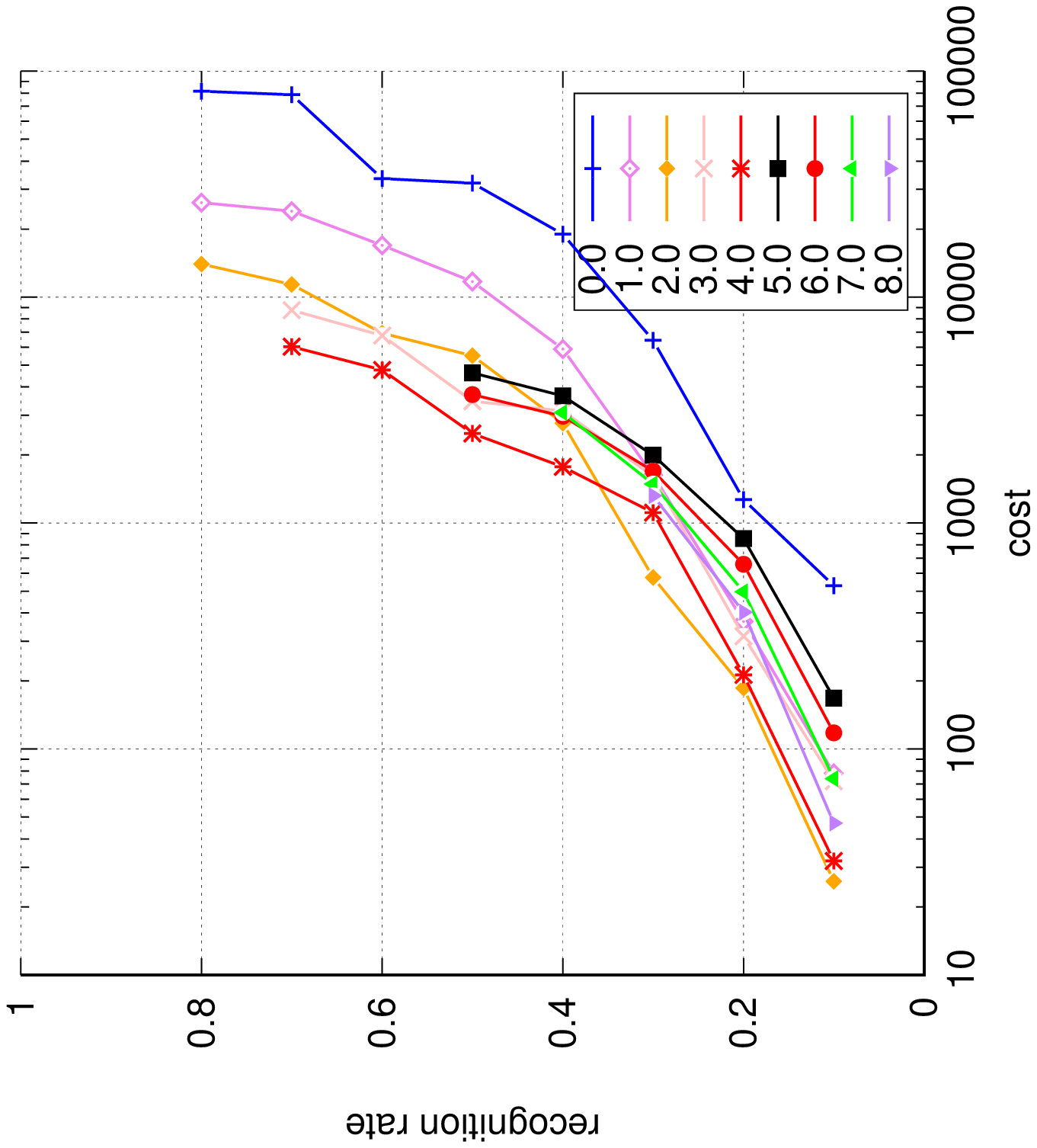}{$frcampus$}{1}%
\hspace*{-1.2cm}\FIGRM{3.4}{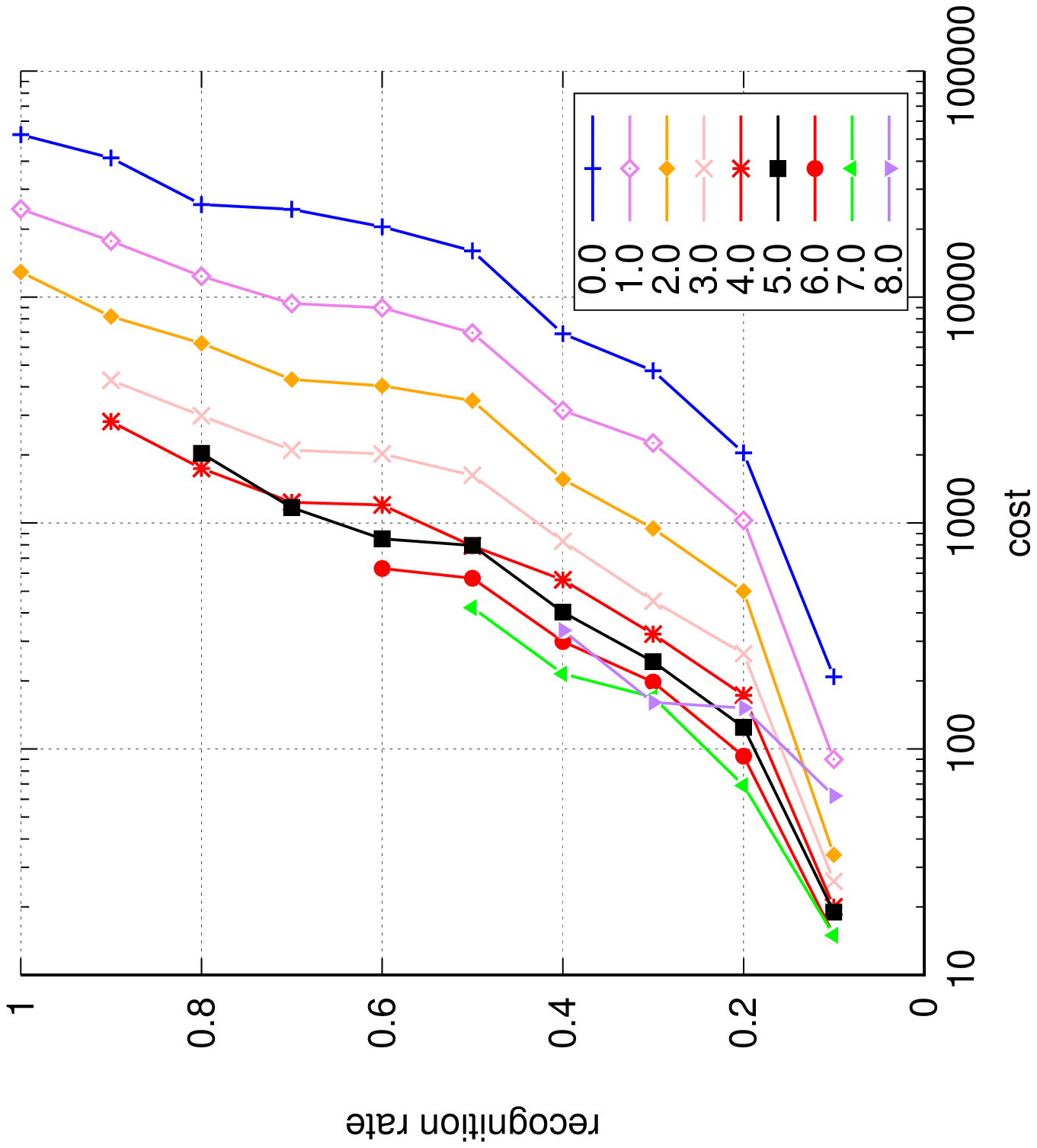}{$mexico$}{1.5}%
\hspace*{-1.2cm}\FIGRM{3.4}{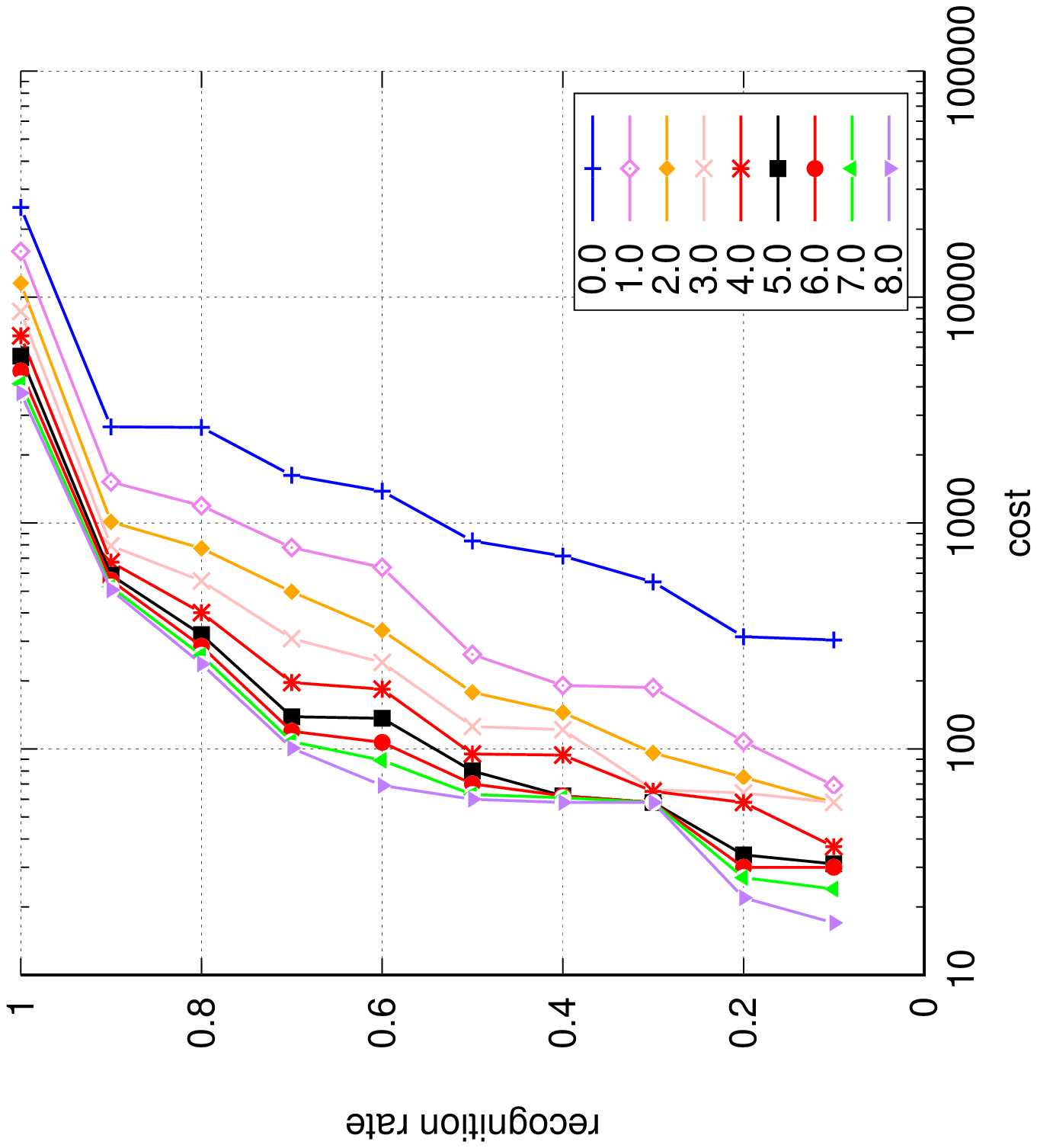}{$seattle$}{1.5}\\
\hspace*{-0.5cm}\FIGRM{3.4}{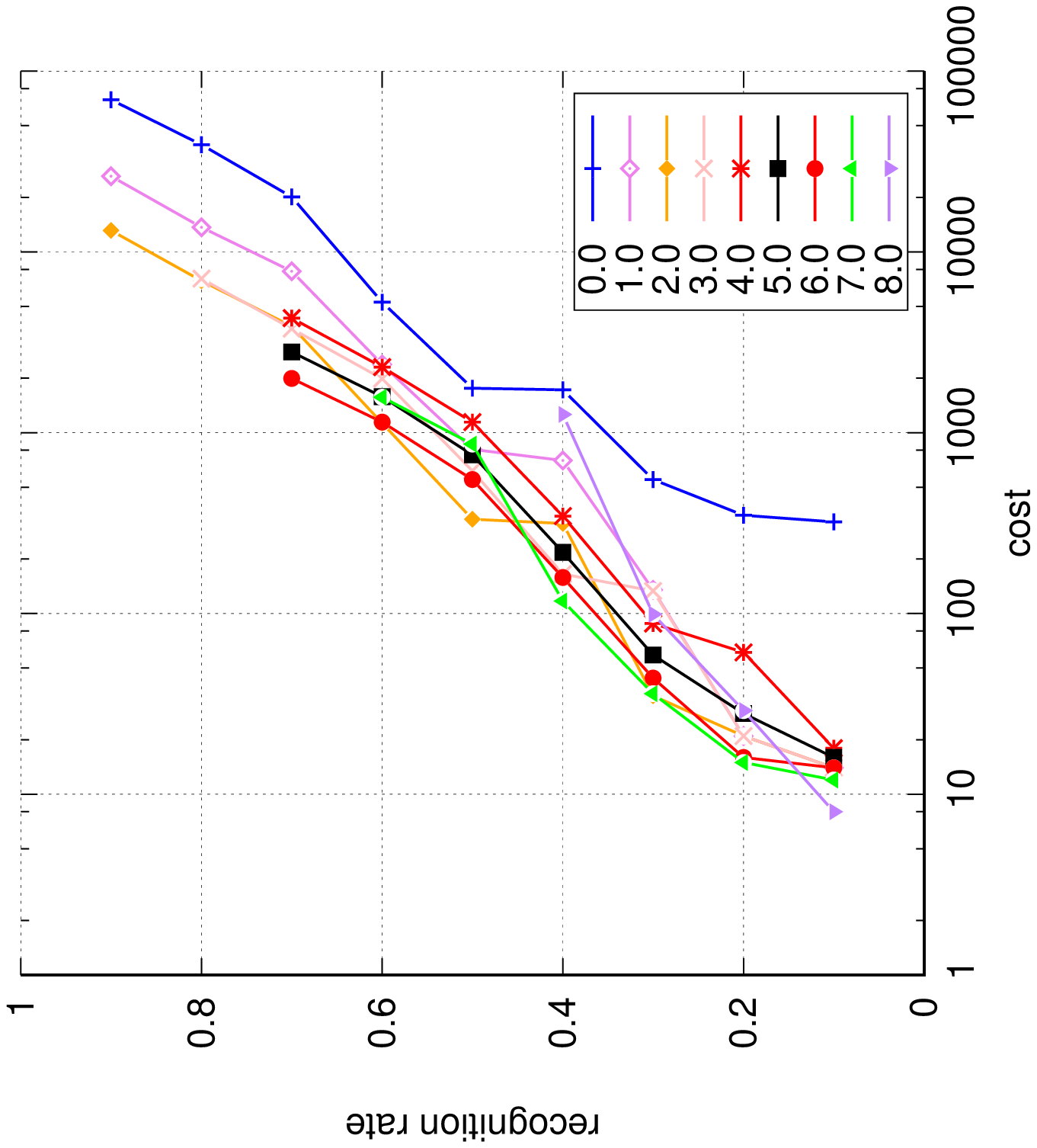}{$edomonton$}{1}%
\hspace*{-1.2cm}\FIGRM{3.4}{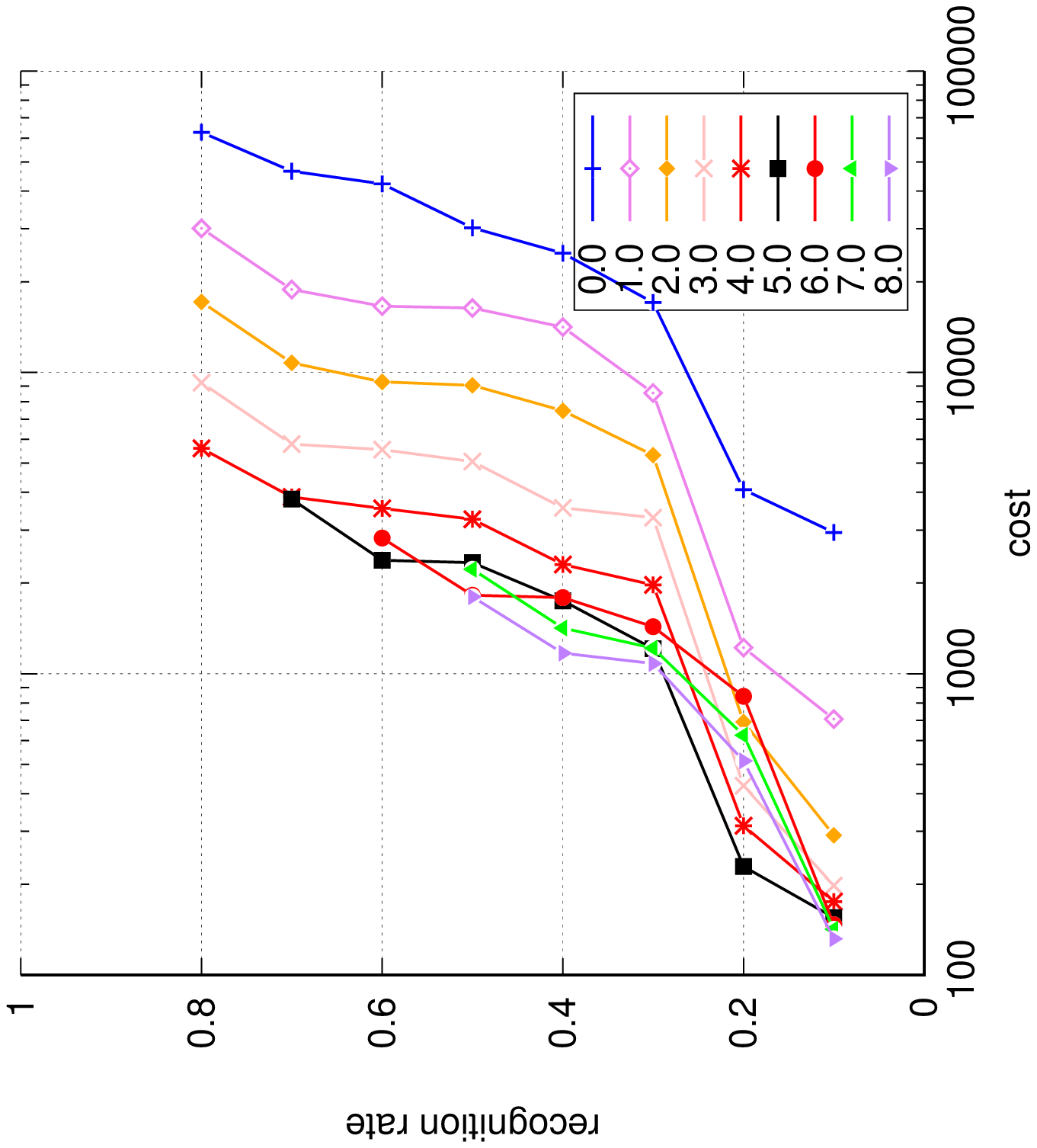}{$albert$}{1}%
\hspace*{-1.2cm}\FIGRM{3.4}{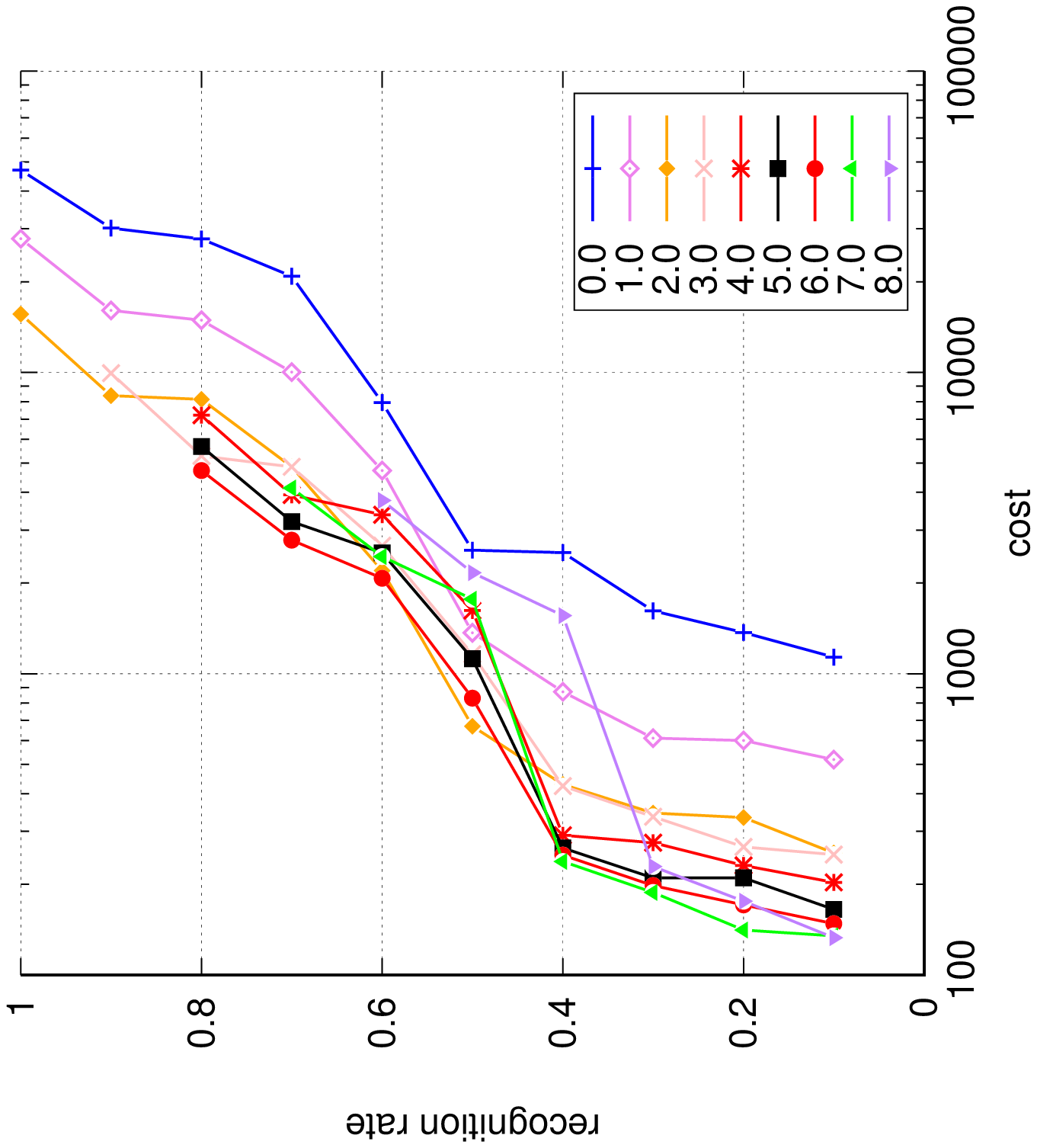}{$csail$}{1}%
\hspace*{-1.2cm}\FIGRM{3.4}{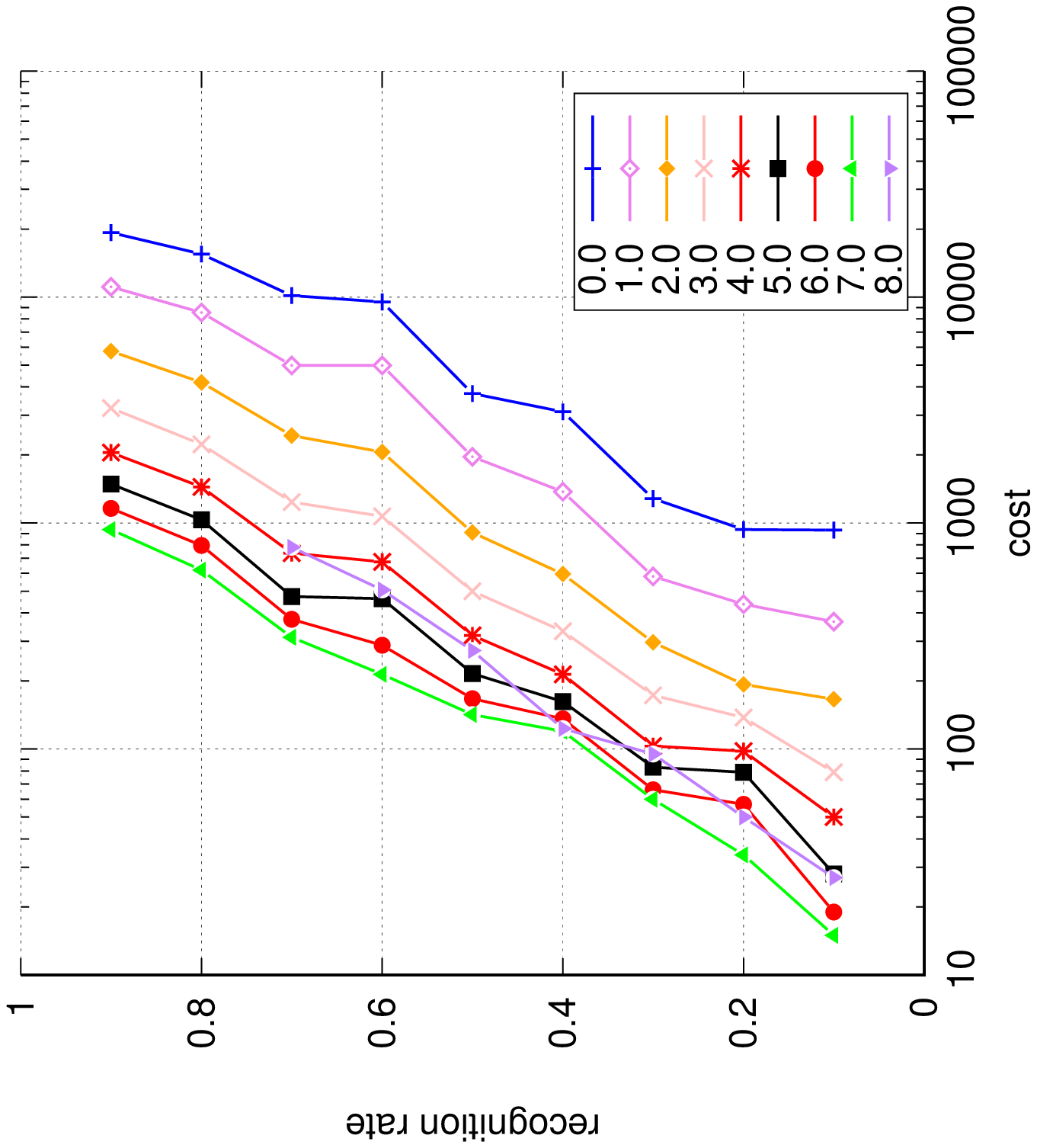}{$orebro$}{1}%
\hspace*{-1.2cm}\FIGRM{3.4}{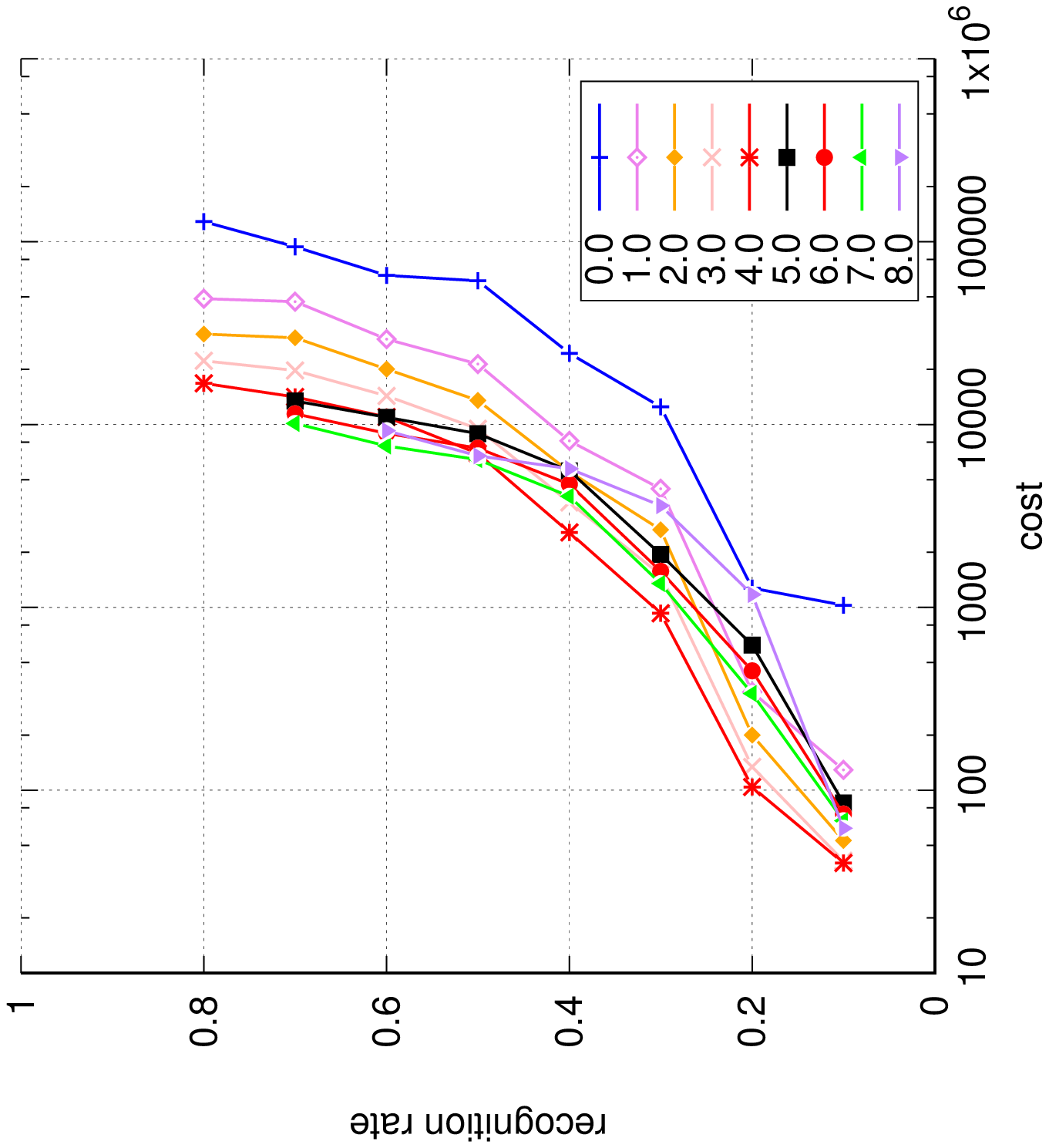}{$fr101$}{0.5}%
\hspace*{-1.2cm}\FIGRM{3.4}{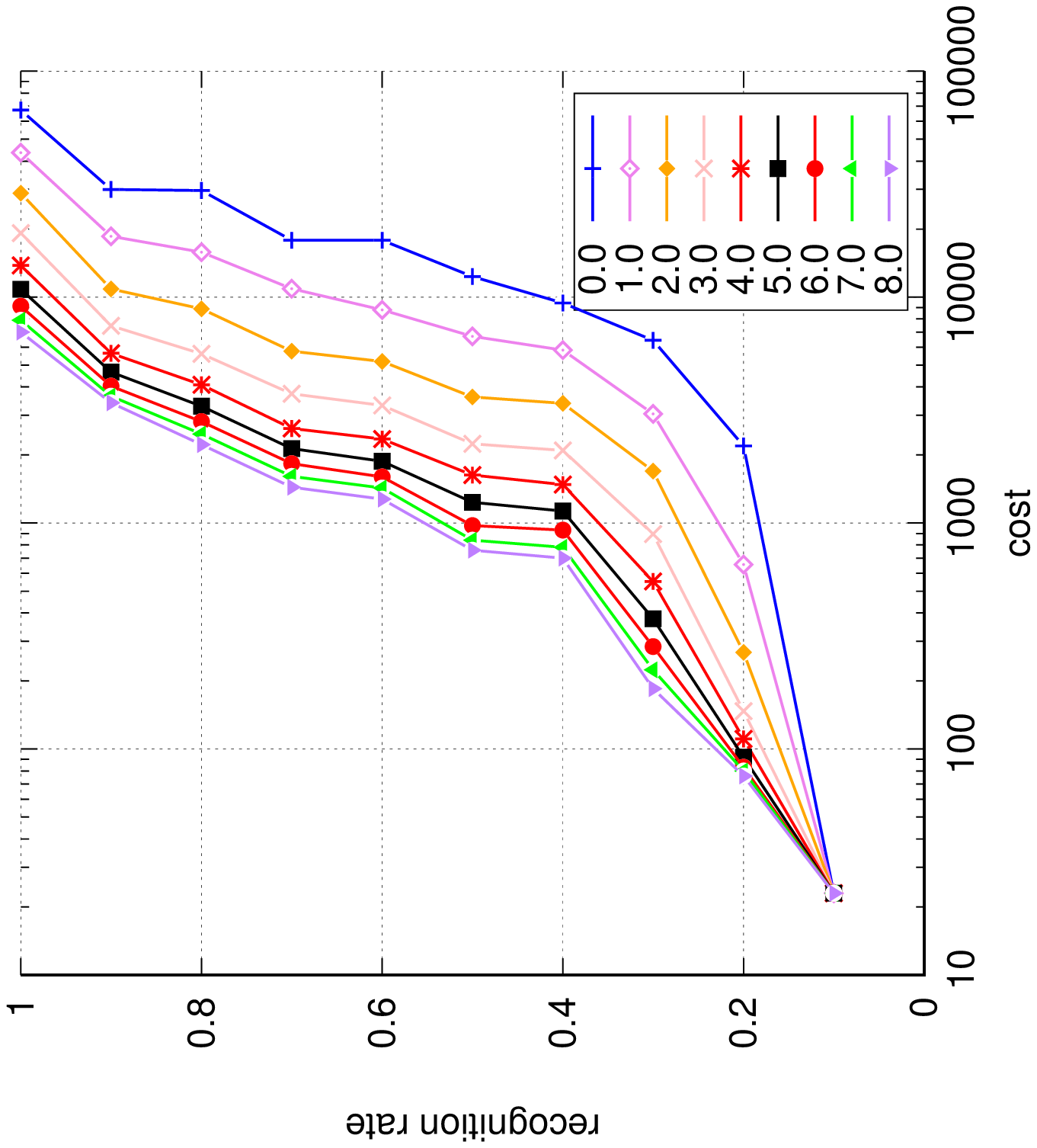}{$aces$}{1.5}%
\hspace*{-1.2cm}\FIGRM{3.4}{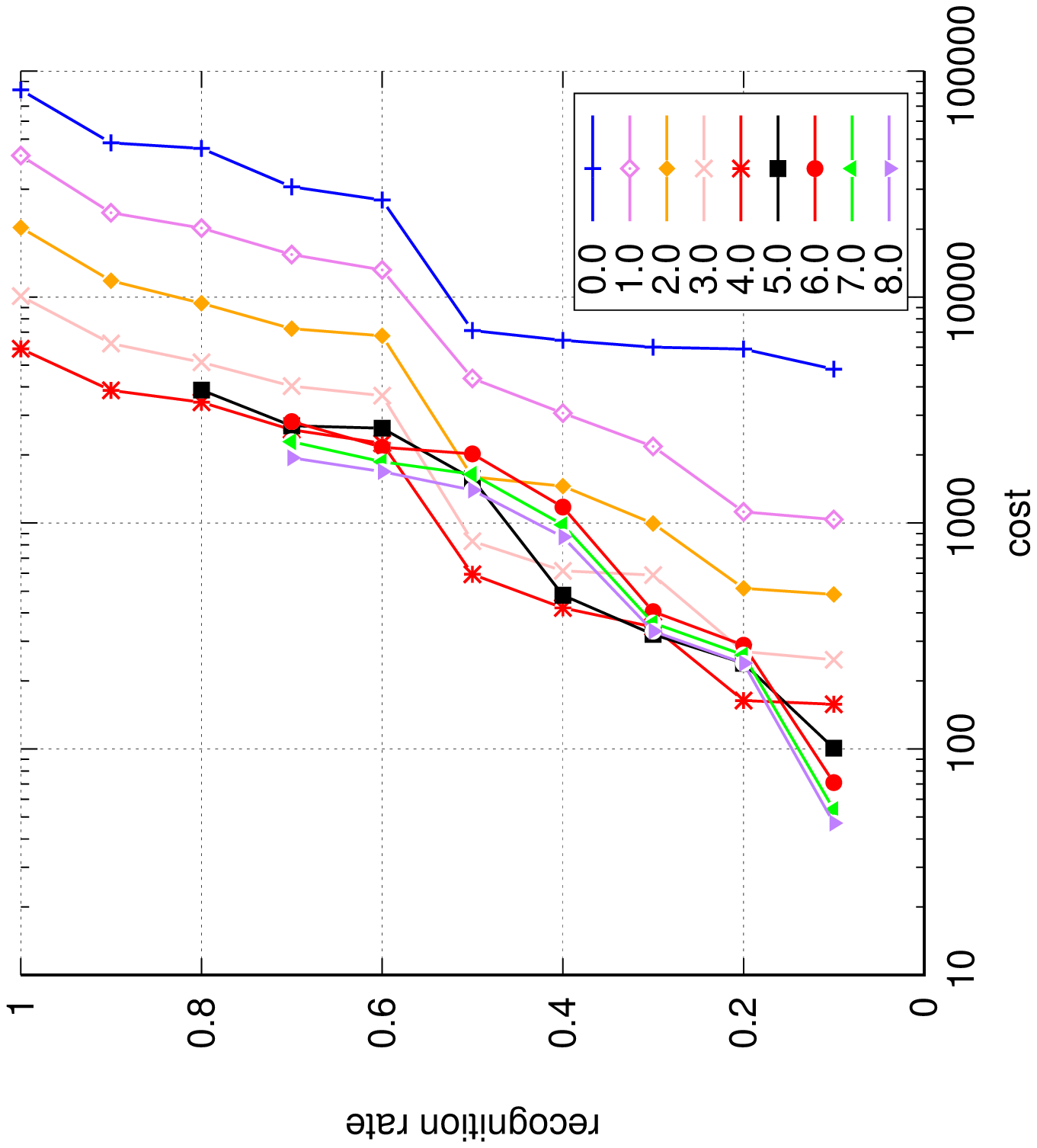}{$fr079$}{0.7}\vspace*{-4mm}\\
\caption{Change detection performance (LMD+CoG, descriptor\#1). 
Horizontal axis: rank of change mask (log-scale). 
Vertical axis: recognition rate.}\label{fig:E}
\end{center}
\vspace*{-6mm}
\end{center}
\end{figure*}
}

\newcommand{\tabB}{
\begin{table}[t]
\caption{Top-$X$\% recognition rate for 2,590 retrieval expewriments }\label{tab:B}
\begin{center}
\vspace*{-5mm}
\begin{tabular}{|r|rrrr|} 
\hline
map retrieval method & Top-10 & & Top-5 & \\ \hline
LMD+CoG & 0.44 & ( 1141 ) & 0.36 & ( 930 ) \\
LMD+CoR & 0.45 & ( 1157 ) & 0.35 & ( 908 ) \\
BoW & 0.25 & ( 645 ) & 0.15 & ( 396 ) \\
LMD+CoR (st) & 0.54 & ( 1393 ) & 0.45 & ( 1161 )	\\
\hline
\end{tabular}
\vspace*{-5mm}
\end{center}
\end{table}
}

\newcommand{\tabA}{
\begin{table}[t]
\caption{ANR performance (in \%, smaller values are better).}\label{tab:A}
\vspace*{-5mm}
\begin{center}
\begin{tabular}{|r|rr|} 
\hline
& ANR per query & ANR per dataset \\
\hline
\hline
1 & 25.8  & 26.1 \\
2 & 25.1 & 26.3 \\
3 & 24.6 & 26.2 \\
4 & 25.6 & 28.7 \\
5 & 25.1 & 26.3 \\
6 & 24.7 & 26.8 \\
7 & 25.1 & 25.6 \\
8 & 25.5 & 26.5 \\
\hline
\end{tabular}
\vspace*{-5mm}
\end{center}
\end{table}
}

\thispagestyle{empty}
\pagestyle{empty}

\title{\LARGE \bf
Local Map Descriptor for Compressive Change Retrieval
}

\author{Tanaka Kanji
\thanks{Our work has been supported in part by 
JSPS KAKENHI 
Grant-in-Aid for Young Scientists (B) 23700229,
and for Scientific Research (C) 26330297.}
\thanks{K. Tanaka is with Faculty of Engineering, University of Fukui, Japan.
{\tt\small tnkknj@u-fukui.ac.jp}}
\thanks{We are grateful to 
Dr. Cyrill Stachniss,
Dr. Giorgio Grisetti,
Dr. Dirk Haehnel,
Dr. Mike Bosse,
Dr. John Leonard,
Dr. Henrik Andreasson, 
Dr. Per Larsson, 
Dr. Tom Duckett,
Dr. Patrick Beeson,
and Dr.  Nick Roy,
for having given access to the datasets used in the paper.}
}

\newcommand{\cofig}[1]{#1} \newcommand{\ocfig}[1]{}

\renewcommand{\cofig}[1]{} \renewcommand{\cofig}[1]{#1}

\maketitle

\begin{abstract}
Change detection, i.e., anomaly detection from local maps built by a mobile robot at multiple different times, is a challenging problem to solve in practice. Most previous work either cannot be applied to scenarios where the size of the map collection is large, or simply assumed that the robot self-location is globally known. 
In this paper, 
we tackle the problem of simultaneous self-localization and change detection, 
by reformulating the problem as a map retrieval problem, and propose a local map descriptor with a compressed bag-of-words (BoW) structure as a scalable solution.
We make the following contributions.
(1)
To enable a direct comparison of the spatial layout of visual features between different local maps, the origin of the local map coordinate 
(termed ``viewpoint") 
is planned by scene parsing
and determined by our ``viewpoint planner"
to be invariant 
against small variations in self-location and changes, 
aiming at providing similar viewpoints for similar scenes (i.e., the relevant map pair). 
(2)
We extend the BoW model to enable the use of not only the appearance (e.g., polestar) but also the spatial layout (e.g., spatial pyramid) of visual features with respect to the planned viewpoint. 
The key observation
is that
the planned viewpoint 
(i.e., the origin of local map coordinate)
acts as a pseudo viewpoint
that is usually required by 
spatial BoW (e.g., SPM)
and 
also by anomaly detection (e.g., NN-d, LOF).
(3)
Experimental results 
on a challenging ``loop-closing" scenario
show that the proposed method 
outperforms previous BoW methods
in self-localization, and furthermore, that 
the use of both appearance and pose information 
in change detection 
produces 
better results than 
the use of 
either information alone.
\end{abstract}

\figA

\section{Introduction}

Change detection, i.e., anomaly detection from local maps built by a mobile robot at multiple different times, is a fundamental problem in robotic mapping and localization \cite{thrun2005probabilistic}, with many important applications ranging from map users (e.g., patrol robots) to mapper robots (e.g., map  maintenance robots). Given a local map of a robot's surroundings as a query, the goal of change detection is to search over a database or a collection of previously built local maps to identify regions that correspond to environment changes 
(e.g., appearance of new objects), 
which comprise the ``change mask". A key issue is that the change mask should not contain ``unimportant" or ``nuisance" forms of change, such as those induced by difference in views and sensor noises. In this sense, change detection is similar in its objectives to anomaly detection \cite{AnomalyDetectionSurvey}, where the goal is to detect anomalies that are interesting to the observer. 

Although the change detection problem has drawn much research attention during the past decade \cite{cvpr13cd,taneja2015geometric,iros07cd,icra13ad,neubert2015superpixel}, it is still a difficult problem to solve in practice. Most previous works either cannot be applied to scenarios where the size of the map collection is large \cite{icra13ad}, or simply assumed that the robot self-location is globally known \cite{taneja2015geometric}. For instance, in \cite{cvpr13cd} and in other related papers, image-based change detection from a cadastral 3D model of a city by using panoramic images captured by a car is addressed under inaccuracies in input geometry, errors in the image's GPS data, as well as, limited amount of information owing to sparse imagery. However, it is not straightforward to extend such approaches to the case of large-scale map collection and globally unknown self-location.
The number of possible local maps 
that need to be examined is 
large and prohibitive without 
{\it compressed} map representation 
and efficient 
map 
{\it retrieval} mechanism.

In this paper, we tackle the problem of 
simultaneous self-localization and change detection, by reformulating the problem as a map retrieval problem, and propose a local map descriptor with 
a compressed bag-of-words (BoW) structure as a scalable solution (Fig. \ref{fig:A}).
We make the following contributions:

(1)
To enable a direct comparison of the spatial layout of visual features between different local maps, the origin of the local map coordinate (termed ``viewpoint") 
is planned by scene parsing
and determined by our ``viewpoint planner"
to be 
{\it invariant}
against 
small variations in
self-location and changes. 
This strategy is inspired by our previous work on grammar-based map parsing \cite{robio11kanji} and unique viewpoint planning \cite{shogo2014m2t}, aiming at providing similar viewpoints for similar scenes (i.e., the relevant map pair). 

(2)
We extend the BoW model to enable the use of not only 
the appearance (e.g., polestar \cite{silani2006star}) 
but also 
the spatial layout
(e.g., spatial pyramid \cite{pyramid2006}) 
of visual features with respect to the 
{\it planned viewpoint}.
The key observation
is that
the planned viewpoint 
(i.e., the origin of local map coordinate)
acts as a pseudo viewpoint
that is usually required by 
spatial BoW (e.g., SPM \cite{pyramid2006})
and 
also by anomaly detection (e.g., NN-d \cite{tax2001one}, LOF \cite{breunig2000lof}).

(3) Experimental results 
on a challenging ``loop-closing" scenario
show that the proposed method 
outperforms previous BoW methods
in self-localization, and furthermore that 
the use of 
{\it both} 
appearance and pose information 
in change detection produces 
better results than 
the use of 
either information alone.
Although our approach is general and can be applicable to various sensor modalities and applications, we focus on an application scenario of 2D pointset maps from laser data \cite{tipaldi2010flirt}, a challenging scenario 
owing to sparse sensing and limited field-of-view.

The use of spatial information 
in the BoW model (e.g., SPM) has been studied in the field of large-scale 
{\it image retrieval} \cite{vocabulary1}. However, to adopt such methods that were originally proposed for image data, we must first determine the viewpoint or the origin of the local map coordinate with respect to which poses of local features are defined, which is a non-trivial task and is our contribution in this paper.

This study is a part of our studies on long-term map learning \cite{icra15a} 
and map-matching \cite{kanji06}, 
and is built on our previous techniques for BoW map retrieval \cite{tanaka2012multi}, grammar based scene parsing \cite{robio11kanji}, unique viewpoint planning \cite{shogo2014m2t}, and change detection \cite{iros97kanji}. However, 
the use of spatial BoW and anomaly detection is not addressed in existing studies.

\figGd

\subsection{Related Work}

Although
various types of ``change detection" tasks (i.e., detecting changes from scenes taken at multiple different times, including video surveillance, remote sensing, medical diagnosis and civil infrastructure) have been studied in the literature \cite{ChangeDetectionSurvey}, in these studies, researchers often 
{\it assume} the availability
of global self-location information (e.g., GPS) or perfect scene registration, and then apply a simple differencing or a more sophisticated method to identify regions of changes. In contrast, our focus is on the applications of robotic mapping and localization \cite{thrun2005probabilistic}, in which the self-location of the robot is globally unknown, and self-localization itself is a challenging topic of ongoing research \cite{VPRsurvey}. 

In recent years, change detection under 
{\it local} uncertainty in viewpoint has also been addressed by several researchers  \cite{cvpr13cd,taneja2015geometric,iros07cd,icra13ad,neubert2015superpixel}. 
In \cite{iros07cd}, a change detection algorithm for ``difference detection" by a patrol robot was considered; however, the focus 
was on NDT-based scene representation and 
its use in change detection 
rather than on viewpoint's uncertainty. 
In \cite{cvpr13cd}, ``city-scale" change detection was built upon the authors' previous work; however, the availability of rough GPS information (i.e., viewpoint)
was assumed.
In \cite{icra13ad}, a reliable solution to change detection from 
LIDAR data was addressed by introducing a method for reasoning on frontiers and occlusions; however, 
the focus was on uncertainty in object location rather than in viewpoint. 
In contrast, we do not rely on prior knowledge of viewpoint, but instead, our viewpoint planner provides an invariant 
viewpoint that acts as a 
{\it pseudo} viewpoint for change detection.

State-of-the-art map matching techniques 
often employ {\it offline} pre-computation 
of an efficient scene model to accelerate online scene retrieval \cite{m3rsm}, rather than employing just online scene matching 
by random sample consensus (RANSAC), 
iterative closest point (ICP), correlation, or other similar techniques such as Chamfer matching. 
Our focus, 
the BoW model, 
is one of the most established approaches for scalable scene retrieval \cite{arandjelovic2012three}. 
Unlike many other map models (e.g., Hough transform) in which a scene is represented by a single global scene descriptor, the BoW model is sufficiently flexible 
to describe a variety of local maps with very different scales 
as unordered collection of local features.

Our focus, pose information of local features, is 
{\it orthogonal} 
to the type of appearance descriptors to be used. 
In \cite{tipaldi2010flirt}, the problem of extracting salient local appearance features from a given local map is discussed. In this study, we employ several different appearance descriptors, derived from our previous studies \cite{kanji09} and \cite{tanaka2012multi}. In general, appearance and spatial information complement each other.

Our approach to anomaly detection,
i.e., detecting previously unobserved patterns in data \cite{AnomalyDetectionSurvey}, 
belongs to the class of 
{\it nearest neighbor (NN) based} 
anomaly detection \cite{tax2001one}, as it is suitable for anomaly detection from very small samples, i.e., a sole relevant pair of query and database maps.

It should be emphasized that our map descriptor does not aim for replacing existing map representations but rather for providing {\it complementary} information. Translation from other map representations (e.g., grid, feature, or view sequence maps) \cite{thrun2005probabilistic} to our map descriptor is often straightforward, which enables efficient BoW map retrieval.

Our viewpoint planner can be viewed as a novel application of {\it grammar-based} scene parsing, which has been previously studied in the fields of point-based geometry, image description, scene reconstruction and scene compression.

\section{Problem}

Our goal is to take a 2D pointset map as a query input, and to search over a size $N$ map database 
to obtain a ranked list of 
self-location candidates (i.e., identifying the database map of change)
and change masks (i.e., anomalies with respect to the database map).

The main steps are as follows (Fig. \ref{fig:G}). 
(1) As a pre-processing step, 
each database map is translated to 
a BoW map descriptor (\ref{sec:lmd}),
and indexed by an inverted file system. 
This pre-processing step
can be done as part of the map-building task. (2) Given a query local map, it is translated to a BoW descriptor and the inverted file is accessed 
based on the word address. 
This visual search process employs 
SPM similarity metric.
As a result, we obtain a ranked list of 
database 
BoW descriptors (\ref{sec:loc}).
(3) Then, we proceed down the list, examining the retrieved database maps one by one and evaluate the anomaly-ness of each feature point from 
each database map (\ref{sec:cd}), 
and if it exceeds a threshold then 
we push back the point's change mask to a list. 
For the results, we obtain a ranked list of change masks.

\section{Approach}\label{sec:app}

The core of our novel map representation,
LMD,
is viewpoint planning (Fig. \ref{fig:F}),
aiming at planning the origin $(x^v, y^v)$ of the local map coordinate,
with respect to which 
poses of local features 
(or pose words)
are defined.

Our approach to viewpoint planning is quite simple: 
determining the ``center" of a given local map 
as its viewpoint (Fig. \ref{fig:A}b). 
This strategy is motivated by an intuition that 
the local map's center is expected 
to be more robust
than such as its boundary.
Our viewpoint planner
first estimates the orientation of the dominant direction 
of the scene structure 
by using the entropy minimization criteria in \cite{olufs2011robust}, 
and defines the $x$-direction of the local map coordinate system 
to be the dominant direction. 
Then, it performs scene parsing 
by using the Manhattan world grammar in \cite{robio11kanji} to extract a set of wall primitives. 
Finally, 
it imposes a grid map 
with an $xy$ resolution of 0.1 m
and assigns each cell ``occupied" (i.e., wall cells), ``unoccupied", 
or ``unknown" label \cite{thrun2005probabilistic}.

The remaining problem is how to determine the center of a given local map. 
Considering that
a key requirement is 
providing similar viewpoints (i.e., ``center") for 
similar scenes (i.e., the relevant map pair) under 
small variations in self-location and changes,
we propose two strategies.
One strategy is to define 
the viewpoint 
as the center of gravity (CoG) of wall cells. Another strategy is 
the so-called center of rooms (CoR), 
which analyzes not the wall primitives 
but the structure of rooms.
More formally,
we analyze the 
rectangular regions of unoccupied cells termed ``rooms" aligned with the orthogonal directions of the quasi-Manhattan world (Fig. \ref{fig:F}). 
We then densely sample rooms and then 
define the viewpoint as the center of gravity of
``dominant" room cells. For robustness, we generate two histograms of 
$f^x(x)$ and $f^y(y)$ of room cells along two dominant directions $x,y$ and consider only those 
dominant cells $(x,y)$ where the histogram values $f^x(x),f^y(y)$ both exceed $90\%$ of the peak values $\max_x f^x(x)$, $\max_y f^y(y)$.

Once viewpoint is planned, it is straightforward
to adopt existing techniques
for spatial BoW (e.g., SPM)
and anomaly detection (e.g., LOF, NN-d),
as explained below.

\subsection{Local Map Descriptor}\label{sec:lmd}

Once the viewpoint is determined,
computation of the BoW vector is straightforward.
We compute 
the local appearance feature descriptors (e.g., polestar, shape context, etc.) 
of local features
as well as their poses with respect to the 
{\it planned}
viewpoint,
and then quantize them to a pose word
$(w_x, w_y)$
and an appearance word $w_a$. 
For the result, we obtain word vectors, each having the form:
\begin{equation}
X = \{ \langle w_a, w_x, w_y \rangle \}.
\end{equation}

We employ an appearance descriptor
that is similar in concept 
to shape context \cite{shapecontext1} 
or polestar \cite{polestar},
which proved successful in our previous work on map matching 
\cite{kanji09}\cite{tanaka2012multi}.
First,
a circular interest region with radius $R$ [m]
is placed at each interest point,
and then,
the interest region is decomposed into 
10 concentric shells 
with radius $r_i$ $(i\in [1, 10])$
and 
$A$ sectors
as
$a_j$ $(j \in [1, A])$
that emerged from the interest point.
To compute an appearance descriptor,
we employ a temporary grid map with spatial resolution 0.01 m.

Each descriptor is mapped to
a $B=10$ bit code
termed visual word
by using a random projection matrix 
derived from our previous work in \cite{iros2011nagasaka}.
This dimension reduction consists of two steps:
(1)
The input $D$-dim descriptor is mapped by a pre-defined 
$B\times D$ random projection matrix to obtain $B$-dim short vector;
(2)
Each element of a $B$-dim short vector
is binarized 
based on its sign
and then the binary vector is encoded to 
a $B$-bit visual word.

\subsection{Global Self-localization}\label{sec:loc}

The global self-localization step 
(i.e., identifying the database map of change)
follows 
the SPM framework \cite{pyramid2006},
by placing a sequence of increasingly coarser grids over the local map and taking a weighted sum of the number of matches that occur at each level of resolution. 
Although the area of a map is variable, we use
three different levels $l=0,1,2$ of 
resolutions
$W2^{-l}$[m]
for a fixed width $W=5$.

Following the pyramid match kernel \cite{pyramid2006}, 
at any resolution $l$, 
the two BoW vectors are compared 
and similarity $I^l$ between a pair of BoW vectors 
is measured
in terms of the number of matched visual words.
The map level similarity $K$ 
between a pair of LMDs
$X$ and $X'$
is defined as:
\begin{equation}
K(X,X')=\frac{1}{2^L} I^0 + \sum_{l=1}^L \frac{1}{2^{L-l+1}} I^l. \label{eqn:pmk}
\end{equation}
We hypothesize and test 
four different orientations $(\pi/2)i$ $(i=0,1,2,3)$
between each pair of quasi-Manhattan aligned query and database maps, 
and use the best hypothesis with the highest similarity score $K$.

Note that each relevant map pair found by the above global self-localization 
is used as input and will be compared by the next change detection step (Fig.\ref{fig:G}).
To accelerate change detection,
we propose to use 
the global self-localization step
to provide a fine alignment of map pairs to 
address the estimation errors of the planned viewpoint.
We compute the mode of a histogram of differences of pose words
\begin{equation}
\Delta x = w_x^{query} - w_x^{DB} ; ~~~ \Delta y = w_y^{query} - w_y^{DB} 
\end{equation}
and use the mode as a temporary offset for the pose words of the database local map during the change detection step.

\subsection{Change Detection}\label{sec:cd}

Given feature locations with respect to the planned viewpoint,
our anomaly detection step
follows 
NN based anomaly detection approaches (LOF, NN-d).
It has two steps:

First, we evaluate LOF \cite{breunig2000lof}
to identify and eliminate 
outlier features,
such as occlusions and noises,
which should not contribute to change detection.
Given a set of feature points, LOF estimates {\it outlier-ness}, 
which is the degree of each point being an outlier.
Let 
$D^{lrd}(p)$
denote the local reachability density of point $p$
\begin{equation}
D^{lrd}(p) = \frac{1}{|S^{NN}(p)|} \sum_{o\in S^{NN}(p)} D^{reach} (p,o),
\end{equation}
where
$S^{NN}(p)$
is the
$K$-NN
of a given point $p$
and
$D^{reach}$
is the 
reachability distance,
which can be seen as an 
extension of 
the point distance function 
to have 
a smoothness effect as defined in \cite{breunig2000lof}.
Then, LOF is evaluated by
\begin{equation}
D^{lof}(p) = \frac{1}{|S^{NN}(p)|} \sum_{o\in S^{NN}(p)} \left[ D^{lrd}(o) / D^{lrd}(p) \right].
\end{equation}
We use a fixed threshold value $T^{lof}=1.6$ \cite{breunig2000lof}
and a point $p$  
is regarded as an outlier
if $D^{lof}(p)$ exceeds $T^{lof}$.

Second, we evaluate NN-d \cite{tax2001one} 
to detect changes 
from appearance words in a query map
by using the appearance words in the database map as training data.
A 60-dim
appearance descriptor (\ref{sec:lmd})
is extracted for each 
``non-outlier" feature point
with a parameter setting of
$(R, A)=(6,10)$,
and then
each element of the vector is binarized
using the average of 
all the 60 elements as a threshold,
which 
yields 
a 60-bit string
as an appearance word for change detection.
Dissimilarity between
a pair of appearance words
is evaluated 
by Hamming distance.
The output of NN-d is an indication of the validity of the object
and 
\begin{equation}
r(p) = | S^{NN}(p)-p |_2 
\Big[ | S^{NN}(p)-S^{NN}(S^{NN}(p)) |_2 \Big]^{-1}.
\end{equation}

We use both the appearance and pose information 
of visual features
for anomaly detection.
We employ the $K=1$ nearest neighbor version of NN-d \cite{tax2001one} 
and thus
$S^{NN}(p)$
is the 1-NN of a given point $p$,
and we search the 1-NN within a fixed size window
$=[-3,3]$[m]$\times$$[-3,3]$[m]
centered at 
a prediction
of the query feature's location,
which is simply
predicted from
the planned viewpoint $p^v$
and 
the original query feature's location
with respect to the query local map's coordinate.
We first 
threshold 
feature points $p_i$ 
by a threshold $r(p_i)>T^{NN-d}$ (e.g., $T^{NN-d}=1.5$)
and then rank 
the surviving feature points
in descending order of
the map level similarity $K$ in (\ref{eqn:pmk}).

Note that 
in our scenario,
the original high dimensional appearance feature vectors are no longer available as they are encoded and replaced 
using a compact BoW model with 60-bit appearance words.
Of course,
this approximation causes
quantization loss.
However,
combining pose and appearance words 
are often
sufficient to obtain accurate results,
as we will see in 
the next experimental section.

\figB

\figF

\section{Experiments}

We conducted 
change retrieval experiments
to verify the efficacy of
the compressed LMD approach. In the following subsections, 
we first describe
the datasets and investigate 
the influence of parameter settings and then report the experimental results on self-localization and change detection.

\subsection{Settings}

We created a large-size map collection from 
a dataset that was kindly provided by researchers
online \footnote{http://www2.informatik.uni-freiburg.de/~stachnis/datasets.html}, which comprises odometry
and laser data logs acquired by a car-like mobile robot in indoor
environments (Fig. \ref{fig:B}). 
The datasets were scan matched to obtain pointset maps from each of 14 datasets
``aces",
``albert",
``belgio",
``csail",
``edomonton",
``fhw",
``fr079",
``fr101",
``frcampus",
``intel",
``mexico",
``mit",
``orebro",
``seattle"
that were obtained from 
the long travel of the mobile robot,
each of which corresponded to 
240-4,800 scans.  
In the current experiments,
each local map corresponds to a 5 m run of the robot,
and
a new local map 
is initialized
every time the robot runs
1 m 
along the trajectory,
yielding around 80-640 occupied cells per map.

For each query map,
we simulated the appearance of new objects 
that were not observed in the database map.
Our change simulator 
places
a change region 
with size 
$w[m]\times h[m]$
$(w,h\in[1.0,2.0])$
at random location
and 
generates a set of small virtual rectangular objects 
within the changed region.
We then run a ray tracing algorithm to 
obtain modified datapoints that 
reflect changed objects.
If a ray hits such a virtual changed object,
the corresponding laser range measurement 
is replaced by the virtual measurement originated from the virtual object.
If any ray does not hit any virtual object,
it is impossible to detect such a change
and thus we re-run the change simulator. 
However,
note that our recognition algorithm does not assume 
any prior knowledge about the shape, pose, 
and size of the environment changes.

We consider a challenging scenario of self-localization,
called loop closing \cite{thrun2005probabilistic},
in which a robot 
traverses a loop-like trajectory and then returns to the previously explored location.
More formally,
the relevant map pair is defined by a database map that satisfies two conditions (light blue lines in Fig. \ref{fig:B}): 
1) The average distance of the datapoints 
from each query datapoint to 
their nearest neighbor database datapoint
is nearer than other candidates.
2) Its distance traveled along the
robot's trajectory is sufficiently distant 
($> 10$ m) 
from that of the query map.
As a result,
a relevant pair of maps
become dissimilar to each other,
which makes our 
self-localization and change detection
tasks
a challenging one.

For each 
of the 14 datasets,
we uniformly sampled
10 
different queries
and 
conducted 
$10\times 14$
$=140$
change detection tasks.
This required
a comparison of
12,320
pairs of query and database maps
in total.

\figC

\subsection{Viewpoint Planning}

Fig. \ref{fig:F}
shows
several examples
of viewpoint planning
using 
the CoR method in \ref{sec:lmd}.
Results from each step of 
the CoR method,
(1) entropy based orientation estimation,
(2) scene parsing by Manhattan world grammar,
(3) sampling of room primitives,
and (4) determining of viewpoint
are shown in the figure.
One can see
that 
our planner frequently provides 
similar viewpoints for the relevant map pair,
even when the robot's trajectories 
are very different between the map pair.

\subsection{Change Detection}

Fig. \ref{fig:C}
shows several examples
of change detection tasks.
Here, 
we used the default value of
the threshold on anomaly-ness
$T^{NN-d}=1.5$.
The figure shows
change detection of
28 
pairs 
uniformly sampled
from the 140 relevant pairs.
It can be seen
that 
change detection
was successful in most 
of the examples shown here.
Exceptionally,
change detection
failed for the cases of
``intel 9, 63",
``fr101 9, 57",
``frcampus 33, 476",
and 
``fr101 19, 48".
In 
``intel 9, 63"
and
``fr101 9, 57",
the change regions 
were near the static large walls;
thus 
the robot could not 
distinguish the changes 
from the known walls.
In 
``frcampus 33, 476",
one can see that
self-localization was not successful
owing to
errors in viewpoint planning,
and as a result,
the robot could not discriminate 
changes from known objects.
In 
``fr101 19, 48",
the robot failed to select
the correct viewpoint orientation
(among the four candidates explained in \ref{sec:loc}),
owing to
unknown objects
caused by changes,
and in consequence,
it failed to detect the changes.
For the other 24 examples,
change detection 
was successful
by using 
both appearance and pose 
information
from our LMD descriptor.

\figD

\figE

\subsection{Performance of Global Self-localization}

This subsection investigates 
the performance of global self-localization.
For the performance evaluation,
we used 
the averaged normalized
rank (ANR) \cite{shogo2013partslam}
as a performance index.
ANR is
a ranking-based performance measure in which a lower value is
better. 
To determine the ANR, we performed 
a number of 
$N$
independent
self-localization tasks with various queries and databases. 
For each task, 
the rank assigned to the ground-truth database map 
by a self-localization method 
of interest was investigated, and the rank was normalized by the
database size $N$. The ANR was subsequently obtained as the average of
the normalized ranks over 
all the self-localization tasks. 
These self-localization tasks 
were conducted using 2,896 different queries and a size 6,193 map
database in total.

Self-localization performance is also measured by its recognition rate.
Given a query local map,
its retrieval result is in the form of 
a ranked list of database maps.
Then,
the recognition rate $y$
is defined
over a set of self-localization tasks,
as 
the ratio $y$ 
of tasks
whose relevant database maps
are correctly included in the top $x$
ranked maps.

\tabB

Fig. \ref{fig:D}
and 
Table \ref{tab:B}
(``LMD+CoG", ``LMD+CoR")
report the results.
It can be seen that 
the proposed LMD method (LMD+CoG, LMD+CoR)
produces good ANR performance (15\%-30\%)
for most datasets considered here.

\subsection{Influence of Appearance Descriptors}

We also 
tested eight different 
appearance feature descriptors
(\ref{sec:lmd})
with different parameter settings of $(R, A)$:
$(1,1)$, 
$(6,3)$, 
$(12,6)$, 
$(1,6)$, 
$(3,6)$, 
$(6,6)$, 
$(3,1)$, 
and $(6,1)$.
The descriptors \#1, \#7 and \#8 can be viewed as variants of 
rotation invariant polestar descriptors \cite{polestar},
whereas the other ones can be viewed as variants of 
the shape context \cite{shapecontext1},
both of which were successful in
our previous work on map matching \cite{tanaka2012multi,kanji09}.
We tested all the eight descriptors
for all the datasets.
For the result,
the ANR per query
was
25.8, 25.1, 24.6, 25.6, 25.1, 24.7, 25.1, and 25.5,
and 
the ANR per dataset
was
26.1, 26.3, 26.2, 28.7, 26.3, 26.8, 25.6, and 26.5
for each descriptor \#1, ..., \#8.
It can be said
the performance difference between the descriptors
was not significant in the current experiments.
For the sake of simplicity,
we employed
descriptor \#1 
(polestar descriptor)
in the following experiments.

\subsection{Comparison against the BoW Method}

To demonstrate the efficacy of the proposed LMD approach,
using spatial information in the BoW model,
we also compared the performance of 
the proposed LMD method
against the previous BoW method
that does not use spatial information,
which is shown in Fig. \ref{fig:D} and Table \ref{tab:B} (``BoW").
As expected,
the performance of the BoW method was
significantly lower.
In contrast to that,
our method performed well as
our method could 
exploit not only 
the appearance 
but also the poses 
of features
to discriminate 
stationary objects (i.e., landmarks)
from changes.

\subsection{Comparison with Stationary Environments}

We also tested the proposed CoR method in 
an alternative scenario of
stationary environments,
i.e., without the appearance of new objects.
Fig. \ref{fig:D}
and Table \ref{tab:B} 
(``LMD+CoR (st)")
show the results,
where the performance 
is much better than in
the case of non-stationary environment, 
(``LMD+CoR").
This results
demonstrate 
the difficulty of 
self-localization 
in non-stationary environments
addressed in this study.

\subsection{%
Performance of Change Detection
}

Fig. \ref{fig:E}
reports the performance of change detection.
To investigate the efficacy of NN-d anomaly detection,
we compared different settings of the threshold on anomaly-ness $T^{NN-d}$:
0.0,
1.0,
...
and 8.0.
As mentioned,
feature points were first thresholded and then ranked
in descending order of map level similarity $K$,
provided by the self-localization step.
When anomaly-ness was ignored
(e.g., $T^{NN-d}=0.0$),
a huge amount of false-positive ``changes" were detected.
When threshold $T^{NN-d}$ was too high
(e.g., $T^{NN-d}=7.0, 8.0$),
many changes were missed.
In consequence,
medium threshold values
$T^{NN-d}=1.0, 2.0, 3.0$
yielded a
good balance
between 
cost and performance.
It can be said that
the use of
not only
pose 
but also
appearance 
information,
provided by the proposed LMD descriptor is important 
for a successful 
self-localization 
and change detection.

\section{Conclusion and Future Work}

In this study,
we tackled and reformulated the problem of simultaneous 
self-localization (i.e., identifying the database map of change)
and 
change detection (i.e., identifying anomalies with respect to the database map)
as a map retrieval problem,
and proposed a local map descriptor 
with a compressed bag-of-words (BoW) structure as a scalable solution. As a primary novelty, the origin (or the viewpoint) 
of the local map coordinate is 
planned by scene parsing
and determined by our ``viewpoint planner"
to be invariant against 
small variations in self-location and changes, 
aiming at providing similar viewpoints for similar scenes 
(i.e., the relevant map pair).
We also extended the BoW model to enable the use of
not only the appearance but also 
the pose of visual features with respect to the planned viewpoint.
The key observation
was that
the planned viewpoint 
(i.e., the origin of local map coordinate)
acts as a pseudo viewpoint
that is usually required by 
spatial BoW (e.g., SPM)
and 
also by anomaly detection (e.g., NN-d, LOF).
Experiments
on a challenging loop-closing scenario
showed
that the proposed LMD method 
outperformed the previous BoW method
in self-localization 
and furthermore that 
the use of both appearance and pose information in change detection produces 
better results than 
the use of either information alone.

Our work is the first significant attempt 
in tapping into 
compressive change retrieval.
While it shows
promises, the problem is far from solved.
In this study,
our experiments were restricted to the 2D environment maps 
with 3 dof viewpoint planning.
An immediate future step is to apply our algorithm on  
3D pointset maps with 6 dof viewpoint planning,
such as from 3D LIDAR data.
Another future direction
is to explore different appearance features.
In this work,
we focused on
shape features
that are effective in pointset maps,
but there are many other types of appearance features
such as color features 
and texture features \cite{icra15a}.
In addition,
we implemented a scene parsing algorithm
using a Manhattan world grammar.
One challenging extension is 
to analyze 
scene structure in more general scenarios
of unstructured environments \cite{shogo2013partslam}.

\bibliographystyle{IEEEtran}
\bibliography{ad}

\end{document}